\documentclass[10pt,twocolumn,letterpaper]{article}

\usepackage[pagenumbers]{cvpr}
\usepackage{pifont}
\usepackage{multirow}
\usepackage[normalem]{ulem}
\useunder{\uline}{\ul}{}

\usepackage{arydshln}

\definecolor{cvprblue}{rgb}{0.21,0.49,0.74}
\usepackage[pagebackref,breaklinks,colorlinks,allcolors=cvprblue]{hyperref}
\usepackage{hyperref}
\def\paperID{23} 
\def\confName{CVPR}
\def\confYear{2025}

\title{Towards Film-Making Production Dialogue, Narration, Monologue Adaptive Moving Dubbing Benchmarks}

\author{
Chaoyi Wang$^{1}$\thanks{Equal contribution} ,\quad Junjie Zheng$^{2}$\footnotemark[1] ,\quad Zihao Chen$^{2}$\footnotemark[1] ,\quad Shiyu Xia$^{2}$ ,\quad Chaofan Ding$^{2}$\\
Xiaohao Zhang$^{2}$ ,\quad Xi Tao$^{2}$ ,\quad Xiaoming He$^{3}$ ,\quad Xinhan Di$^{4}$\thanks{Corresponding author}\\
$^{1}$Shanghai Institute of Microsystem and Information Technology, CAS, China,\\
$^{2}$AI Lab, Giant Network, China,
$^{3}$Fudan University, China \\
{\tt\small chaoyiwang@mail.sim.ac.cn, zhengjunjie,chenzihao,xiashiyu, dingchaofan, zhangxiaohao@ztgame.com,}\\
{\tt\small v-taoxi@ztgame.com, 22210700113@m.fudan.edu.cn, dixinhan@ztgame.com}
}

\begin{document}
\maketitle
\begin{abstract}
Movie dubbing has advanced significantly, yet assessing the real-world effectiveness of these models remains challenging. A comprehensive evaluation benchmark is crucial for two key reasons: 1) Existing metrics fail to fully capture the complexities of dialogue, narration, monologue, and actor adaptability in movie dubbing. 2) A practical evaluation system should offer valuable insights to improve movie dubbing quality and advancement in film production. To this end, we introduce Talking Adaptive Dubbing Benchmarks (TA-Dubbing), designed to improve film production by adapting to dialogue, narration, monologue, and actors in movie dubbing. TA-Dubbing offers several key advantages: 1) Comprehensive Dimensions: TA-Dubbing covers a variety of dimensions of movie dubbing, incorporating metric evaluations for both movie understanding and speech generation. 2) Versatile Benchmarking: TA-Dubbing is designed to evaluate state-of-the-art movie dubbing models and advanced multi-modal large language models. 3) Full Open-Sourcing: We fully open-source TA-Dubbing at \url{https://github.com/woka-0a/DeepDubber-V1} including all video suits, evaluation methods, annotations. We also continuously integrate new movie dubbing models into the TA-Dubbing leaderboard at \url{https://github.com/woka-0a/DeepDubber-V1} to drive forward the field of movie dubbing.
\end{abstract}    
\section{Introduction}
Dubbing involves adding the correct human voice to a video's dialogue, ensuring synchronization with the character's lip movements, and conveying the emotions of the scene. It plays a vital role in film, television, animation and gaming, enhancing immersion and effectively conveying emotions and atmosphere. Existing dubbing methods can be categorized into two groups, both of which focus on learning different styles of key prior information to generate high-quality voices. The first group focuses on learning effective speaker style representations \cite{chen2022v2c,hassid2022more,wan2018generalized,cong2023learning}. The second group aims to learn appropriate prosody by utilizing visual information from the given video input \cite{cong2023learning,hu2021neural,lee2023imaginary,zhao2024mcdubber}. the accuracy of these priors is insufficient and inadequate for movie dubbing in real-world scenarios. For example, adaptive dubbing for different types, such as dialogue, narration, and monologue, as well as fine-grained attributes such as expected ages and genders, has not been thoroughly studied \cite{cong2024styledubbermultiscalestylelearning,hu2021neural}. With the rapid advancement of large language reasoning models with step-by-step thinking ability \cite{o1-min,o1,GPT-o3-mini,grok-3,claude-3-7,deepseekai2025deepseekr1incentivizingreasoningcapability,QwQ-32B} and methods that enhance reasoning capabilities to interpret visual information through CoT, MLLM have increasingly shown their potential in multi-modal reasoning and understanding tasks \cite{penamakuri2024AudiopediaAudioQA,Insight,cheng2024videollama2advancingspatialtemporal,kim2024videoiclconfidencebasediterativeincontext,bonomo2025visualragexpandingmllm,liu2024EnhancingVisualReasoning,li2025imaginereasoningspacemultimodal,sahili2024FairCoTEnhancingFairness,zheng2024ThinkingLookingImproving}.
However, Both groups fail to meet the practical demands of real-world scenes. With the growth of movie dubbing models, there is an increasing need for effective evaluation frameworks to support moving dubbing in the film making.   

However, existing metrics for movie dubbing such as SPK-SIM, WER, MCD and MCD-SL \cite{lee2023imaginary,zhao2024mcdubber} are insufficient for film production needs. Because they focus solely on non-adaptive evaluation of speech generation based on several dubbing settings \cite{yemini2024lipvoicergeneratingspeechsilent,jang2024facesspeakjointlysynthesising}. As in real-world scenarios, industrial-level movie dubbing, both in film production and post-production stages, is expected to adapt to various factors. Two key factors are the adaptability of actors and the dialogue, narration, and monologue. Therefore, we propose TA-Dubbing, a comprehensive benchmark suite for evaluating video clip comprehension and the quality of dubbing adaptable to dialogue, narration, and monologue, with the goal of producing high-quality speech towards film production.

\begin{figure*}[t!]
    \centering
    \resizebox{\linewidth}{!}{
    \includegraphics[]{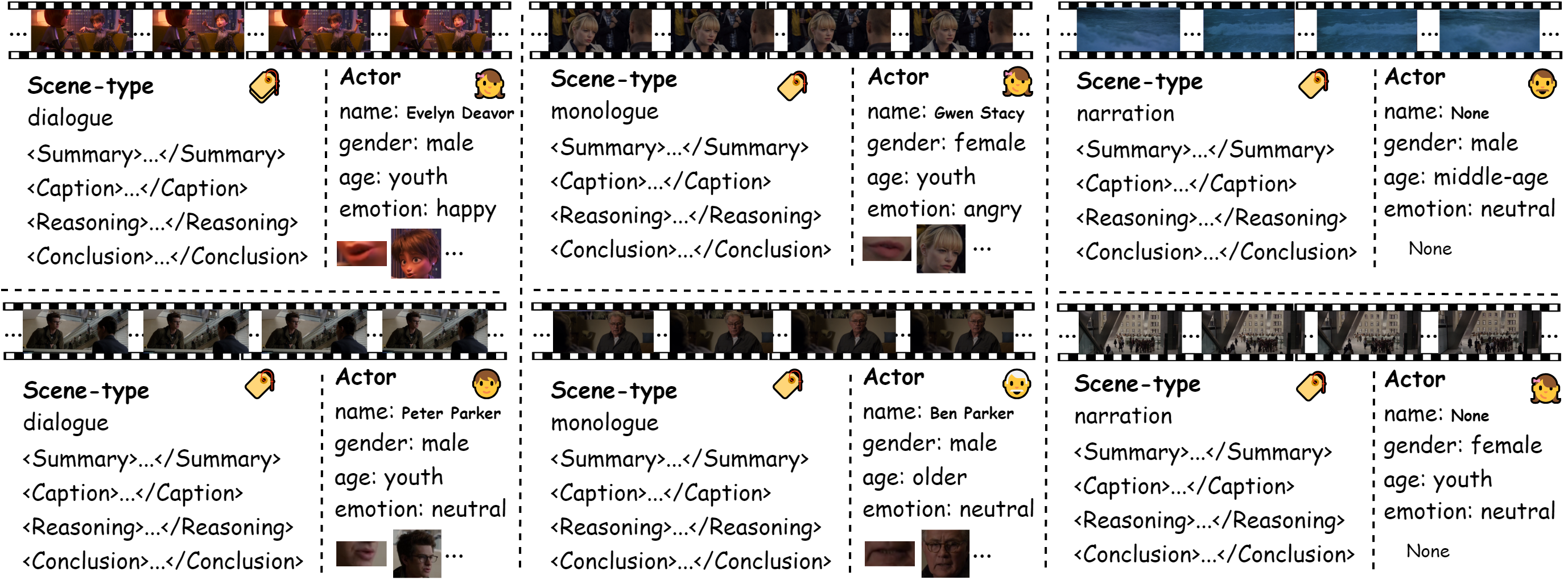}
    }
    \caption{Proposed dataset with multi-type annotations, including annotation for lips, faces, scene-type, actor name, actor gender, actor age, voice emotion and chain-of-thoughts annotations for dialogue, narration and monologue.}
    \label{fig:dataset}
\end{figure*}
\section{Related Work}

\subsection{Visual Voice Cloning} Recent dubbing approaches integrate visual and textual cues to enhance speech-video alignment and emotional richness \cite{congLearningDubMovies2023,hu2021neural,lee2023imaginary}. Some methods center on robust speaker identity for multi-speaker scenarios \cite{cong2024styledubbermultiscalestylelearning,zhang2024from}, such as injecting pretrained speaker embeddings into phoneme encoders and mel-spectrogram decoders \cite{zhang2024from}. Others focus on visual representations to boost prosody, for example, HPMDubbing \cite{congLearningDubMovies2023} fuses lip motion, facial expressions, and scene content, while MCDubber \cite{zhao2024mcdubber} extends modeling to adjacent video contexts for smoother transitions. Despite these advancements, challenges such as the adaptability of actors and the dialogue, narration, and monologue—two of the most important factors in film production—are not sufficiently evaluated.     



\subsection{Evaluation of Movie Dubbing Models}
Existing movie dubbing models typically use metrics like SPK-SIM, WER, MCD and MCD-SL \cite{cong2024styledubbermultiscalestylelearning,zhang2024from,sahipjohnDubWiseVideoGuidedSpeech2024} for the evaluation of the speech generation. Despite the typical evaluation of generated speech, which focuses solely on non-adaptive movie dubbing with fixed settings \cite{zhangSpeakerDubberMovie2024,congStyleDubberMultiScaleStyle2024,cong2024emodubberhighqualityemotion}. The current evaluation system falls short of meeting the requirements for assessing movie dubbing in film production. One of the key challenges in evaluation is the adaptivity of movie dubbing to actors, dialogue, narration, and monologue. Encouraged via rapid development of reasoning ability of visual reasoning combines perception and cognition, evaluated in tasks like VQA \cite{90339209302102270} and visual entailment \cite{Song2022CLIPMA}. Besides, with large language models offering step-by-step reasoning \cite{GPT-o3-mini}, vision-language systems can better interpret visual scenes \cite{liu2023visual}. We propose TA-Dubbing to evaluate both the comprehension of movie clips (particularly the recognition of dialogue, narration, monologue, and actors) and the corresponding assessment of adaptive movie dubbing. Additionally, we provide corresponding step-by-step annotations to support adaptive movie dubbing for film production.                 

\begin{figure*}[t!]
    \centering
    \resizebox{\linewidth}{!}{
    \includegraphics[]{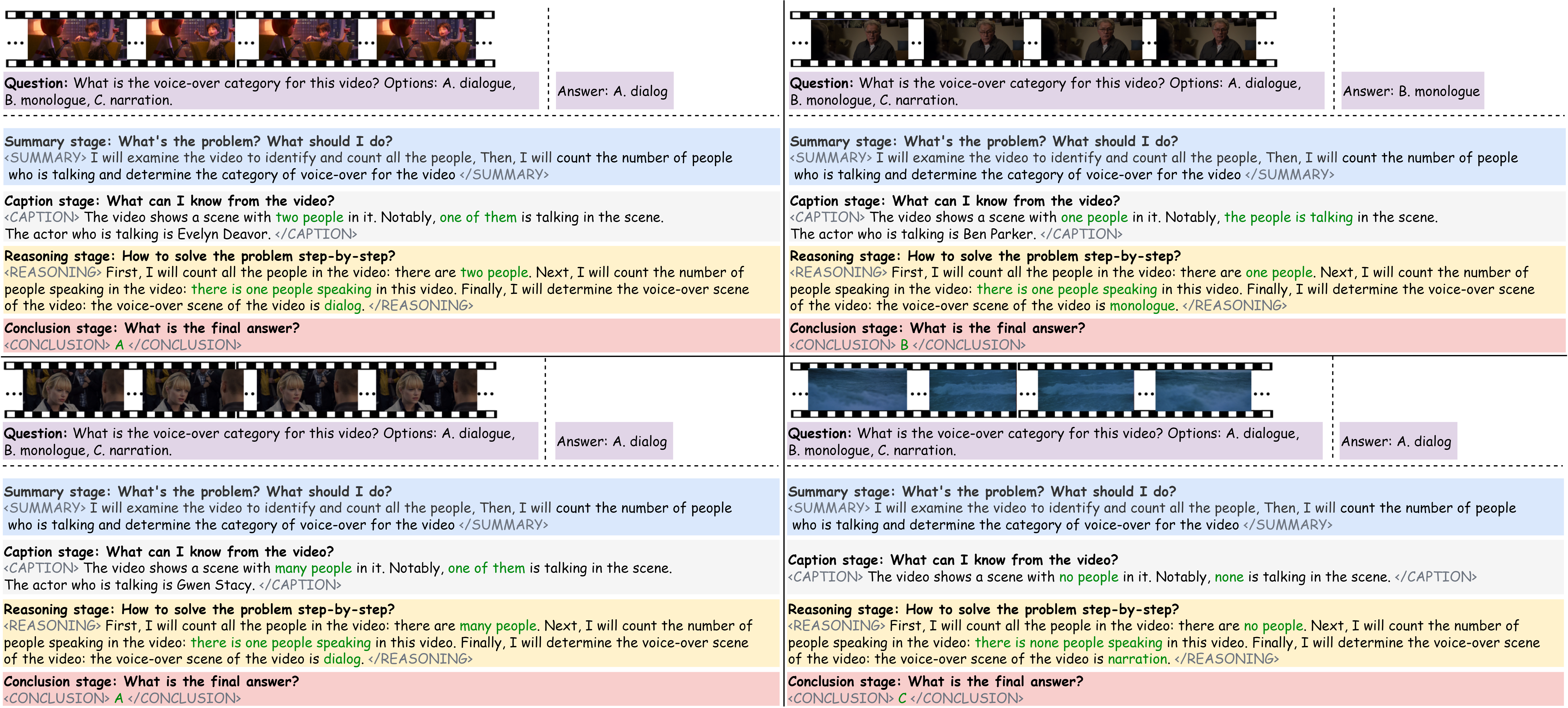}
    }
    \caption{The reasoning stages of movie scene type(dialogue, narration, monologue) CoT annotations.}
    \label{fig:dataset_cot}
\end{figure*}

\begin{table*}[t!]
\centering
\caption{
\textbf{Towards Adaptive(Dialogue, Narration, Monologue) Moving Dubbing Evaluation Results Per Dimension}
For speech generation setting, we use the target speaker's speech as voice prompt if the predict scene type is correct and use random speaker's speech as voice prompt if the predict scene type is not correct.}
\resizebox{\linewidth}{!}{
    \begin{tabular}{cc cccc cccc}
    \toprule
    \multicolumn{2}{c}{\multirow{4.5}{*}{Models Name}} & \multicolumn{4}{c}{Scores on dialogue(A), monologue(B) and narration(c)} & \multicolumn{4}{c}{Speech Generation} \\ \cmidrule(r){3-6} \cmidrule(r){7-10}
    \multicolumn{2}{c}{} & \multicolumn{1}{l}{Ave.Pre(\%) $\uparrow$} & \multicolumn{1}{l}{A.Pre(\%) $\uparrow$} & \multicolumn{1}{l}{B.Pre(\%) $\uparrow$} & \multicolumn{1}{l}{C.Pre(\%) $\uparrow$} &  &  &  & \\
    \multicolumn{2}{c}{} & \multicolumn{1}{l}{Ave.Recall(\%) $\uparrow$} & \multicolumn{1}{l}{A.Recall(\%) $\uparrow$} & \multicolumn{1}{l}{B.Recall(\%) $\uparrow$} & \multicolumn{1}{l}{C.Recall(\%) $\uparrow$} & SPK-SIM(\%) $\uparrow$ & WER(\%) $\downarrow$ & MCD $\downarrow$ & MCD-SL $\downarrow$ \\ 
    \multicolumn{2}{c}{} & Ave.F1-Score(\%) $\uparrow$ & A.F1-Score(\%) $\uparrow$ & B.F1-Score(\%) $\uparrow$ & C.F1-Score(\%) $\uparrow$ &  &  &  &  \\ \midrule
    \multicolumn{2}{c}{} & 100.00 & 100.00 & 100.00 & 100.00 &  &  &  &  \\
    \multicolumn{2}{c}{GT} & 100.00 & 100.00 & 100.00 & 100.00 & 100.00 & 17.38 & 0.00 & 0.00 \\
    \multicolumn{2}{c}{} & 100.00 & 100.00 & 100.00 & 100.00 &  &  &  &  \\\midrule
    \multicolumn{10}{c}{MLLMs based} \\ \midrule
    \multirow{2}{*}{} &  & 59.66 & 88.06 & 54.55 & 36.36 &  &  &  &  \\
    \multirow{2}{*}{} & Zero-shot & 12.8 & 26.22 & 6.38 & 5.80 & - & - & - & - \\
    \multirow{2}{*}{GPT-4o \cite{GPT-4o}} &  & 20.61 & 40.41 & 11.43 & 10.00 &  &  &  &  \\ \cdashline{2-10}
    \multirow{2}{*}{} &  & 68.66 & 84.71 & 77.78 & 43.48 &  &  &  &  \\
     & Finetune & 64.59 & 91.11 & 44.68 & 57.97 & - & - & - & - \\ 
     \multirow{2}{*}{} &  & 64.74 & 87.79 & 56.76 & 49.69 &  &  &  &  \\ \midrule
    \multicolumn{10}{c}{Dubbing Models} \\ \midrule
     \multicolumn{2}{c}{HPMDubbing \cite{congLearningDubMovies2023}} & - & - & - & - & 61.06 & 199.40 & 8.82 & 11.88 \\
     \multicolumn{2}{c}{Speaker2Dub \cite{zhang2024from}} & - & - & - & - & 61.73 & 84.42 & 8.75 & 10.78 \\
     \multicolumn{2}{c}{StyleDubber \cite{cong2024styledubbermultiscalestylelearning}} & - & - & - & - & 64.03 & 52.69 & 8.62 & 8.89 \\ \bottomrule 
    \end{tabular}
}
\label{table:scene_and_speech_benchmark}
\end{table*}

\section{Benchmark Suite}

\subsection{Movie Dubbing Dataset Suite}
We have created a $200$ multimodal CoT movie dubbing dataset to generate adaptive and high-quality movie dubbing. This dataset contains $140k$ video clips. Based on CoT reasoning and CoT-like guidance \cite{xu2025llavacotletvisionlanguage}, we utilize a professional annotation team to label the following dataset. We develop a CoT reasoning framework to guide subsequent movie dubbing tasks, as illustrated in Figures \ref{fig:dataset} and \ref{fig:dataset_cot}. Specifically, a step-by-step instruction process with video input is designed to enable efficient and accurate movie scene type recognition(dialogue, narration, monologue).
As shown in Figure \ref{fig:dataset_cot}, \texttt{<SUMMARY>}\texttt{</SUMMARY>} provides a high-level overview of the entire scene, while \texttt{<CAPTION>}\texttt{</CAPTION>} describes the characters in the video. During the \texttt{<REASONING>}\texttt{</REASONING>} stage, the reasoning process is divided into five steps:
\begin{itemize}
    \item \makebox[1.1cm][l]{\textbf{Step 1.}} Count the numbers of people in the video.
    \item \makebox[1.05cm][l]{\textbf{Step 2.}} Distinguish whether the people in the video are talking or not.
    \item \makebox[1.1cm][l]{\textbf{Step 3.}} Recognize the faces of the actors.
    \item \makebox[1.1cm][l]{\textbf{Step 4.}} Distinguish whether the movie contains dialogue, narration, or monologue.
    \item \makebox[1.11cm][l]{\textbf{Step 5.}} Conclusion and give the answer.
\end{itemize}

And then \texttt{<CONCLUSION>}\texttt{</CONCLUSION>} stage give the final answer. Each stage is initiated at the model’s discretion, without external prompt engineering frameworks or additional prompting. Specifically, we provide the model with four pairs of special tags, these tags correspond to summarizing the response approach, describing relevant image content, conducting reasoning, and preparing a final answer, respectively. The proposed dataset consists of $130k$ video clips for training and $10k$ video clips for testing. 

\subsection{Evaluation Metrics Suite}
To comprehensively assess the performance of our system, we report a set of recognition and speech quality metrics. The recognition performance is measured through Precision, Recall, and F1 Score, both as overall averages and on a per-class basis. In addition, the generated speech quality is evaluated by assessing speaker similarity and acoustic fidelity.

\paragraph{1. Dialogue, Narration, Monologue Recognition Evaluation Metrics} 
For evaluating the classification performance, we compute the following metrics:
\begin{itemize}
    \item \textbf{Precision}:In our evaluation, we report overall Precision (Ave.Pre) as well as the accuracy for each individual category(Dialogue, Narration, Monologue) (A.Pre, B.Pre, C.Pre) to better understand both the general performance and the class-specific behavior of the model. 
    \item \textbf{Recall}: We report both an overall recall (Ave.Recall) and the recall for each class (A.Recall, B.Recall, C.Recall), which helps assess the performance across different categories.
    \item \textbf{F1 Score}:We present the overall F1 Score (Ave.F1-Score) along with the F1 Score for each class (A.F1-Score, B.F1-Score, C.F1-Score) to offer a comprehensive view of the classifier's performance.
\end{itemize}

\paragraph{2. Speech Quality Evaluation Metrics}  
To evaluate the quality of the generated speech, we utilize the following metrics:
\begin{itemize}
    \item \textbf{Speaker Similarity (SPK-SIM)}: Uses a pre-trained speaker encoder to compute the cosine similarity between the generated speech and a reference speech sample. Higher percentages indicate greater timbre consistency \cite{cong2024styledubbermultiscalestylelearning}.
    \item \textbf{Word Error Rate (WER)}: Computed by transcribing the generated speech using Whisper-V3 and comparing it to the reference text. A lower WER signifies higher pronunciation accuracy \cite{radford2022robustspeechrecognitionlargescale}.
    \item \textbf{Mel Cepstral Distortion (MCD)} and \textbf{MCD-SL}: MCD measures the spectral differences between the generated and reference speech using Dynamic Time Warping (DTW), while MCD-SL further incorporates differences in speech duration. Lower values indicate a closer match in spectral characteristics and temporal alignment \cite{battenberg2020locationrelativeattentionmechanismsrobust}.
\end{itemize}

\paragraph{3. Actor Recognition Evaluation Metrics}
To evaluate the quality of the actor recognition, similarly, we utilize evaluation metrics: precision, recall and f1 score.
\begin{itemize}
    \item \textbf{Precision}: The accuracy for the attribute(the name, the gender, the age, the emotion) of the actor are calculated and the average precision is calculated.
    \item \textbf{Recall}: The recall for the attribute(the name, the gender, the age, the emotion) of the actor are calculated and the average recall is calculated.
    \item \textbf{F1 Score}: The f1 score for the attribute(the name, the gender, the age, the emotion) of the actor are calculated and the average fa score is calculated.
\end{itemize}

The expected performance direction is indicated: higher percentages for Precision, Recall, F1-Score and SPK-SIM denote better performance, whereas lower percentages for WER, MCD, and MCD-SL denote improved results.


\section{Experiments}
We initially evaluate both the state-of-the-art movie dubbing models for the evaluation of speech generation and the understanding of movie on the recognition of dialogue, narration, monologue, actor which both driving forward film-making production towards adaptive actors and dialogue, narration and monologue.  

\noindent\textbf{Dialogue, Narration, Monologue Evaluation.}
We initially conduct evaluation on the recognition of dialogue, narration and monologue for GPT4o\cite{GPT-4o}. The average Precision is $68.66\%$, the precision of dialogue, narration and monologue is $84.71\%$, $77.78\%$ and $43.48\%$.
Similar evaluation is demonstrated in Table \ref{table:scene_and_speech_benchmark}.

\noindent\textbf{Speech Generation Evaluation.}
We initially conduct evaluation on the quality of speech. Particularly, the $SPK-SIM$ is from $61.06$ to $64.03$, the $WER$ is from $52.69$ to $199.40$, the $MCD$ is from $8.62$ to $8.82$, the MCD-SL is from $8.89$ to $11.88$ for a variety of state-of-the-art movie dubbing models \cite{congLearningDubMovies2023,zhang2024from,cong2024styledubbermultiscalestylelearning}. 

\noindent\textbf{Actor Attribute Evaluation.}
To be noted, according to our initial experiments, the performance of GPT4o\cite{GPT-4o} for the recognition (precision, recall and f1 score) is low. We will conduct experiments on other state-of-the-art multi-modal large language models.   

\begin{figure}[t!]
    \centering
    \resizebox{\linewidth}{!}{
    \includegraphics[]{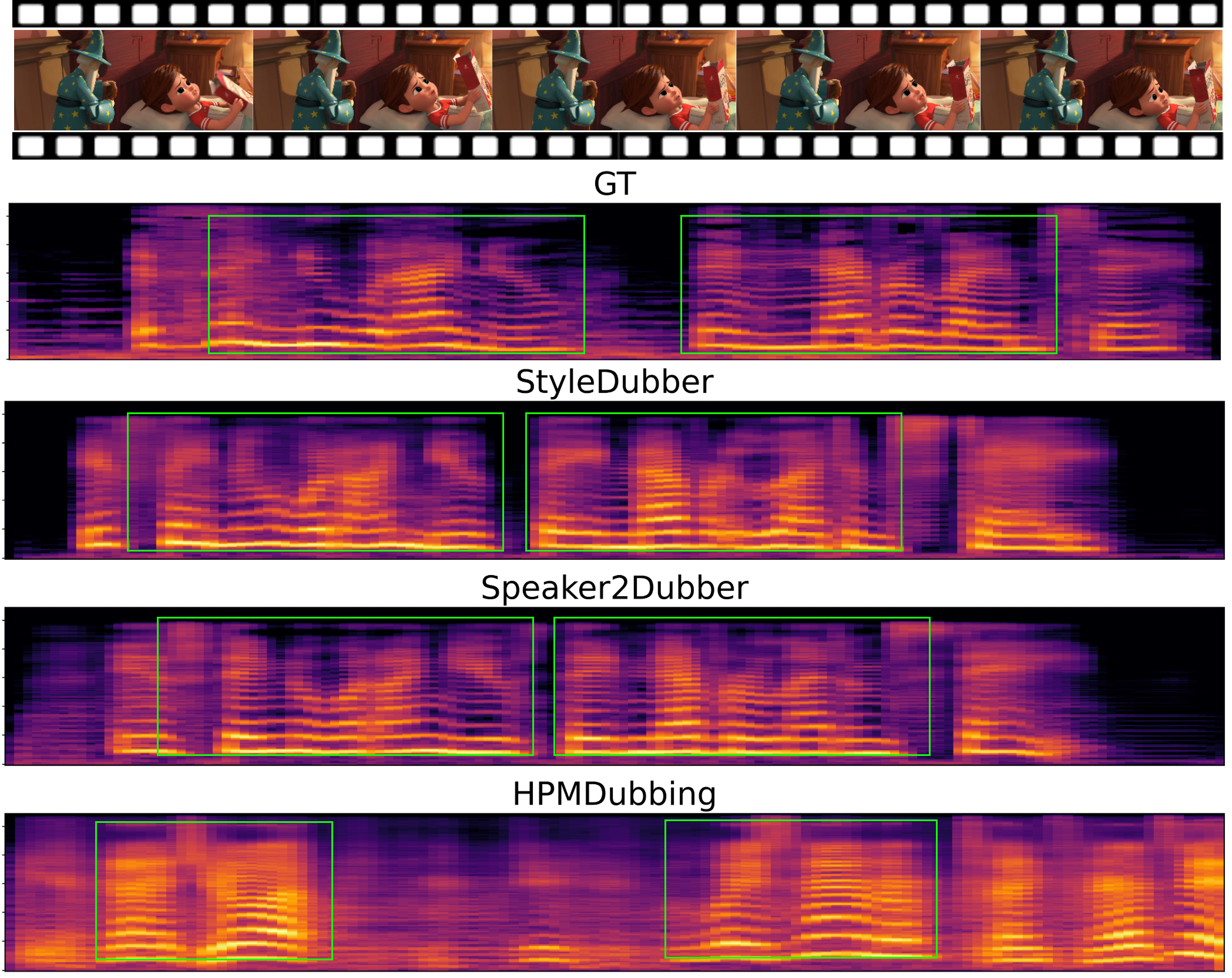}
    }
    \caption{Visualization of generated movie dubbing samples via state-of-the-art models. The green rectangles highlight key regions that have significant differences in overall expressiveness.}
    \label{fig:mel_visualization}
\end{figure}

\section{Discussion}
With the ongoing development of movie dubbing, a comprehensive evaluation of these models is essential to assess current advancements and support movie dubbing in film production. In this work, We take the first step forward by proposing TA-Dubbing, a comprehensive benchmark suite for evaluating movie dubbing models, focusing on two adaptive factors: dialogue, narration, monologue, and actors.     

{
    \small
    \bibliographystyle{ieeenat_fullname}
    \bibliography{main}
}

\end{document}